\documentclass[letterpaper]{article} 
\usepackage{aaai2026}  
\usepackage{times}  
\usepackage{helvet}  
\usepackage{courier}  
\usepackage[hyphens]{url}  
\usepackage{graphicx} 
\usepackage{natbib}  
\usepackage{caption} 
\usepackage{algorithm}
\usepackage{algorithmic}
\usepackage{multirow} 
\usepackage{booktabs}
\usepackage{amsmath}
\usepackage{amssymb}
\usepackage[table]{xcolor}
\usepackage{nicematrix}

\usepackage{newfloat}
\usepackage{listings}
\DeclareCaptionStyle{ruled}{labelfont=normalfont,labelsep=colon,strut=off} 
\floatstyle{ruled}
\newfloat{listing}{tb}{lst}{}
\floatname{listing}{Listing}
\title{GS-Checker: Tampering Localization for 3D Gaussian Splatting}
\author {
    Haoliang Han\textsuperscript{\rm 1},
    Ziyuan Luo\textsuperscript{\rm 1},
    Jun Qi\textsuperscript{\rm 1},
    Anderson Rocha\textsuperscript{\rm 2},
    Renjie Wan\textsuperscript{\rm 1}\thanks{Corresponding author. This work was carried out at the Renjie Group, Hong Kong Baptist University.}
}
\affiliations {
    \textsuperscript{\rm 1}Department of Computer Science, Hong Kong Baptist University\\
    \textsuperscript{\rm 2}Institute of Computing, University of Campinas\\
    \{haolianghan, ziyuanluo\}@life.hkbu.edu.hk, tsinghua.qijun@gmail.com, arrocha@unicamp.br, renjiewan@hkbu.edu.hk
}
\usepackage{bibentry}

\begin{document}

\maketitle

\begin{abstract}
Recent advances in editing technologies for 3D Gaussian Splatting (3DGS) have made it simple to manipulate 3D scenes. However, these technologies raise concerns about potential malicious manipulation of 3D content.
To avoid such malicious applications, localizing tampered regions becomes crucial. In this paper, we propose \textbf{\textit{GS-Checker}}, a novel method for locating tampered areas in 3DGS models.
Our approach integrates a 3D tampering attribute into the 3D Gaussian parameters to indicate whether the Gaussian has been tampered.
Additionally, we design a 3D contrastive mechanism by comparing the similarity of key attributes between 3D Gaussians to seek tampering cues at 3D level. 
Furthermore, we introduce a cyclic optimization strategy to refine the 3D tampering attribute, enabling more accurate tampering localization. Notably, our approach does not require expensive 3D labels for supervision. 
Extensive experimental results demonstrate the effectiveness of our proposed method to locate the tampered 3DGS area.
\end{abstract}

\begin{links}
    \link{Code}{https://github.com/haolianghan/GS-Checker}
\end{links}

\section{Introduction}
Methods for editing 3D Gaussian Splatting (3DGS)~\cite{3dgs} have gained significant attention. Despite the impressive capabilities of these advanced editing techniques~\cite{chen2024gaussianeditor,chen2024dge,wu2024gaussctrl}, they could be exploited to alter 3DGS content maliciously and misuse tampered models. Therefore, detecting such tampering is essential to prevent malicious applications.

\begin{figure}[t]
  \centering
  \includegraphics[width=\linewidth]{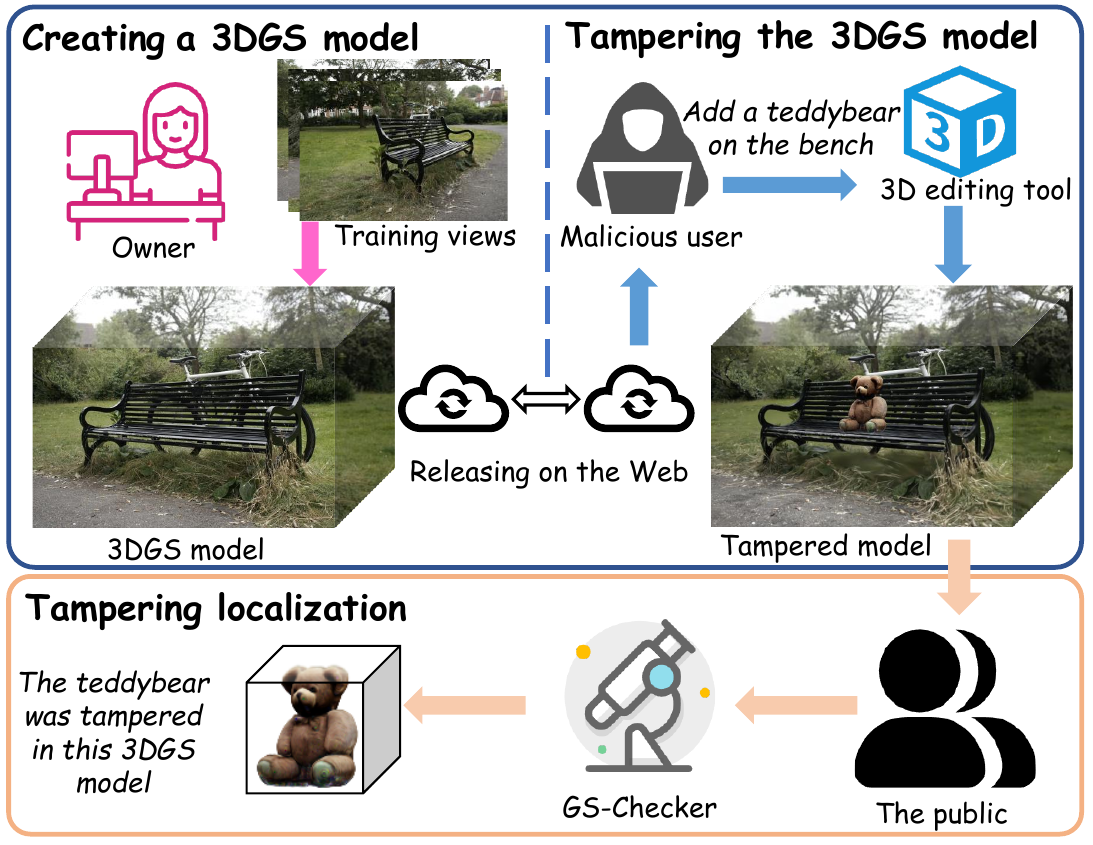}
  \caption{Our proposed scenario for 3DGS tampering localization. When the owner creates a 3DGS model and releases it publicly over the web, malicious users can manipulate it using advanced 3D editing tools for illegal purposes. Upon the release of a tampered 3DGS, the public can use our proposed GS-Checker to identify the tampered areas, thereby protecting the integrity and ownership of the 3DGS model.}
  \label{fig:short}
\end{figure}

In our scenario, illustrated in Figure~\ref{fig:short}, a malicious user uses advanced editing tools to alter specific properties of a 3DGS model originally created by its owner. This tampering raises two main concerns: \textbf{1)} it may undermine the original owner's rights over the 3DGS model, and \textbf{2)} the tampered model could convey meanings with potentially negative societal impacts. Our objective is to detect such unauthorized modifications, safeguarding the integrity and rightful ownership of 3DGS models.

A straightforward approach is to leverage established tampering localization mechanisms. While tampering localization has been well-researched for decades, most approaches~\cite{dong2022mvss,guo2023hierarchical,guillaro2023trufor,zhang2024editguard,ma2023iml} focus primarily on 2D data like images or video frames. When applied to 3DGS~\cite{3dgs}, these mechanisms can only detect tampering in rendered images.
However, without considering 3D spatial correlations across viewpoints, 2D methods struggle to accurately locate tampered areas in each viewpoint.
For example, as shown in Figure~\ref{fig:authentic}, directly applying the 2D image tampering localization approach can mistakenly identify authentic areas as tampered. 

A major reason for the above limitations is that it overlooks the specific challenge in our scenario: \textbf{\textit{the tampering occurs within the 3DGS model}}. 2D tampering localization~\cite{dong2022mvss,wang2022objectformer,guo2023hierarchical,guillaro2023trufor,zhang2024editguard,ma2023iml} can only partially identify tampering traces from rendered 2D images, and these traces do not accurately represent the tampering operations in 3D. We need a 3D tampering localization method capable of directly detecting tampering at the model level. Then, regardless of the viewpoint rendered, we can identify the tampered areas.

To achieve this goal, we propose \textbf{\textit{GS-Checker}}, a framework that can conduct the tampering localization for the 3DGS model.
In traditional 2D methods, tampered pixels often deviate from normal statistical distributions, revealing structural or contextual inconsistencies. 
This phenomenon also extends to the 3D domain, where tampered 3D Gaussians exhibit certain anomalies in their parameter distributions. 
These anomalies reflect differences between tampered and authentic 3D Gaussians, providing critical cues for identifying tampering. 
By analyzing these deviations, we can effectively detect the tampering trace across the entire 3D representation.
In our whole design, we fully make use of the parameters within 3D Gaussians to identify the tampering trace. 
Specifically, we introduce a \emph{3D contrastive mechanism} that compares the similarity of key attributes between 3D Gaussians to seek tampering cues at the 3D level.  
To better exploit the properties of the 3DGS model, a \emph{unique tampering attribute} is integrated into each 3D Gaussian parameters.
This tampering attribute is initialized by extracting tampering cues at the 2D level and projecting them into 3D. 
Besides, we further introduce a \emph{cyclic optimization strategy} to update the 3D tampering attribute by projecting it into 2D.
This allows us to optimize at both 2D and 3D levels jointly, resulting in precise localization of 3DGS tampering.

\begin{figure*}
  \centering
  \includegraphics[width=\linewidth]{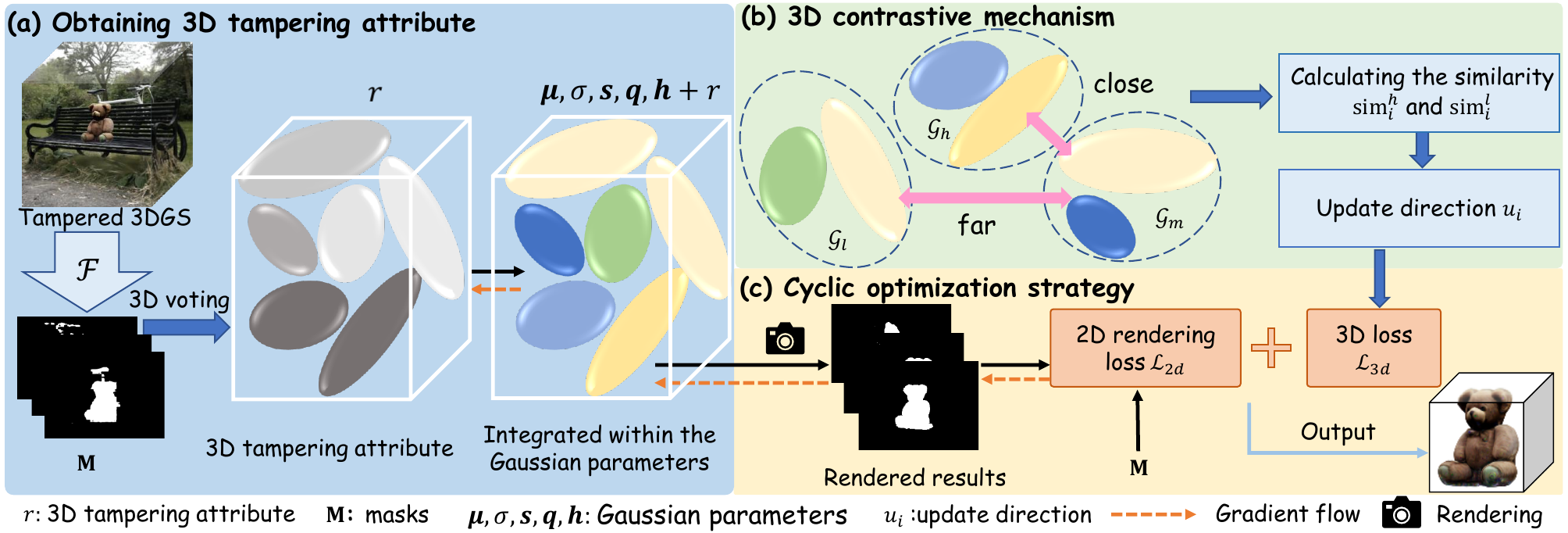}
  \caption{Illustration of our proposed method. First, the \emph{3D tampering attribute} is integrated into the 3D Gaussian parameters and initialized via 3D voting. Next, a \emph{3D contrastive mechanism} is introduced to seek tampering cues by comparing the similarity of key attributes between 3D Gaussians. Finally, a \emph{cyclic optimization strategy} is employed to iteratively refine the tampering attribute by projecting it back into the 2D space, enabling joint optimization across both 2D and 3D levels.}
  \label{fig:framework}
\end{figure*}

Figure~\ref{fig:framework} illustrates our framework. 
The tampering attribute is integrated within the 3D Gaussians parameters, which is initialized via 3D voting.
The 3D contrastive mechanism seeks tampering cues at 3D level by comparing the similarity of key attributes between 3D Gaussians.
The cyclic optimization strategy updates the tampering attribute iteratively by rendering it back into 2D.
Our approach fully exploits the unique properties of 3D Gaussians, performing joint optimization across 2D and 3D levels to precisely localize tampered regions in 3DGS models.
To sum up, our key contribution can be concluded as follows:
\begin{itemize}
    \item A pioneering method, \textbf{\textit{GS-Checker}}, for locating manipulated areas in the 3DGS model by leveraging 3D Gaussian properties.
    \item A 3D contrastive mechanism that seeks tampering cues by analyzing the similarity of key attributes among 3D Gaussians.
    \item A cyclic optimization strategy to refine the 3D tampering attribute for precise localization.
\end{itemize}

Our GS-Checker operates without requiring label-intensive 3D annotations for supervision. We evaluate GS-Checker under various settings, and the results demonstrate its effectiveness.

\section{Related work}
\noindent\textbf{3D editing.} In recent years, 3D technology has made significant progress~\cite{liu2021editing,xu2022deforming,chen2024gaussianeditor,zhang2024gs,song2025align,li2024variational,li2025modeling}. 
Some approaches~\cite{bao2023sine,gao2023textdeformer,wang2022clip,wang2023nerf} leverage the CLIP model~\cite{radford2021learning} to facilitate editing using text prompts or reference images.
Moreover, some approaches~\cite{chen2024gaussianeditor,chen2024dge,wu2024gaussctrl} develop editing techniques for 3DGS models, which can effectively avoid the shortcomings of slow speed and limited control of NeRF-based methods.
For instance, GaussianEditor~\cite{chen2024gaussianeditor} leverages Gaussian semantic tracing and Hierarchical Gaussian Splatting (HGS) for precise and stable 3D editing, and designs a specialized 3D inpainting algorithm to streamline object removal and integration.
With these methods, malicious users may easily use them to manipulate 3DGS models for negative applications. This motivates us to develop tamper localization techniques for 3DGS model, thus avoiding these 3D editing methods from being used for malicious purposes.\\
\noindent\textbf{Tampering localization.} Recently, AI security technologies~\cite{zhang2024editguard,song2024protecting,huang2025marksplatter,song2024geometry,luo2025nerf,huang2024gaussianmarker} have made some progress.
Early image forensic techniques primarily aim at particular types of manipulations~\cite{islam2020doa,li2018fast,li2019localization}. 
Recently, some general tamper localization methods also endeavor to detect artifacts and anomalies within manipulated images~\cite{dong2022mvss, guillaro2023trufor, ma2023iml,zhang2025omniguard}.
For example, MVSS-Net~\cite{dong2022mvss} leverages multi-view feature learning and multi-scale supervision to simultaneously exploit boundary artifacts and the noise perspective of images.
SAFIRE~\cite{kwon2025safire} employs point prompting to segment forged image regions, allowing for the partitioning of images into multiple source regions and naturally focusing on the uniform characteristics within each region.
IML-ViT~\cite{ma2023iml} builds a ViT-based image manipulation localization model with high-resolution capacity, multi-scale feature extraction, and manipulation edge supervision.
However, these methods are unable to achieve the tampering localization for 3DGS. 

\section{Preliminary}
\noindent\textbf{3D Gaussian Splatting.}
3DGS~\cite{3dgs} represents an advanced technique for modeling 3D scenes. Beginning with a sparse set of points derived from Structure-from-Motion (SfM)~\cite{snavely2006photo}, the primary aim of 3DGS is to refine these points into a set of 3D Gaussians to achieve high-quality novel view synthesis. The scene is constructed as a collection of 3D Gaussians:
\begin{equation}
    \mathcal{G}(x) = e^{-\frac{1}{2}(x-\mu)^T \Sigma^{-1} (x-\mu)}.
\end{equation}
In this equation, $x$ represents any position in the 3D scene, $\mu$ is the mean position of the 3D Gaussian, and $\Sigma$ is the covariance matrix of the 3D Gaussian. By utilizing a scaling matrix $S$ and a rotation matrix $R$, the covariance matrix $\Sigma = RS S^T R^T$ can be derived, ensuring its positive semi-definiteness. These 3D Gaussians are then projected onto 2D Gaussians for rendering through volume splatting. During the rendering process, 3DGS employs a conventional neural point-based method~\cite{kopanas2021point,kopanas2022neural} to calculate the pixel color $C$ by blending $N$ depth-ordered points:
\begin{equation}
    C = \sum_{i \in N} c_i \alpha_i \prod_{j =1}^{i-1} (1 - \alpha_j),
\end{equation}
where $c_i$ denotes the color estimated by the spherical harmonics (SH) coefficients of each Gaussian, and $\alpha_i$ is determined by evaluating a 2D Gaussian with covariance \( \Sigma' \), multiplied by a per-point opacity.

\noindent\textbf{3DGS editing.} Recent advancements have elevated 2D diffusion processes to 3D, extensively applying these methods in 3D editing. 
One way~\cite{mikaeili2023sked,poole2022dreamfusion,sella2023vox} involves feeding the noised rendering of the current 3D model and other conditions into a 2D diffusion model~\cite{rombach2022high}, using the generated scores to guide model updates. The other way~\cite{raj2023dreambooth3d,shao2023control4d,chen2024it3d} focuses on 2D editing based on given prompts for multiview rendering of a 3D model, creating a multi-view 2D image dataset to guide the 3D model. GaussianEditor~\cite{chen2024gaussianeditor} leverages the exemplary properties of 3DGS's explicit representation to enhance 3D editing, and employs the aforementioned guidance methods. It can be formulated as:
\begin{equation}
    \mathcal{L}_{\text{edit}} = G(\Theta ; p, e),
\end{equation}
where $G$ denotes the guidance, $p$ represents the rendered camera pose, $e$ represents the prompt, and $\Theta$ represents the 3DGS model parameters. 

\noindent\textbf{Our scenario.}
As shown in Figure~\ref{fig:short}, when a 3DGS model is created by its owner and publicly distributed over the web, malicious users can manipulate it by using advanced 3D editing tools and utilize it for some illegal applications. 
To prevent 3DGS model from being used for malicious purposes, the public need a method to detect such manipulation. 
Upon release of a tampered 3DGS model, the public can employ some approaches to locate which areas have been tampered, thus protecting the integrity and rightful ownership of the 3DGS model.
To achieve this, we propose \textbf{\textit{GS-Checker}}, a method for locating tampered areas in the 3DGS model.

\section{Proposed GS-Checker}
In this section, we present GS-Checker in a comprehensive manner.
As illustrated in Figure~\ref{fig:framework}, our GS-Checker framework incorporates several key mechanisms to locate tampered areas in the 3DGS model. 
First, the tampering attribute is integrated into the parameters of each 3D Gaussian and initialized using a 3D voting method.
Next, a 3D contrastive mechanism is introduced to assess the similarity of key attributes between 3D Gaussians, seeking tampering clues in 3D space.
Furthermore, a cyclic optimization strategy is implemented to update the tampering attribute iteratively by rendering it into 2D. Our method can perform optimization at 2D and 3D levels jointly, resulting in accurate outcomes.
\subsection{Obtaining 3D tampering attribute}
In our approach, the attribute is initially obtained at 2D level and then project back to the 3D space to affiliate with each 3D Gaussian. 
Specifically, we directly feed rendered images from the 3DGS model into a pretrained 2D tampering localization network to obtain the masks as follows:
\begin{equation}
\mathbf{M}_{j}=\mathcal{F}(\mathcal{R}_{j}) ,
\end{equation}
where $\mathcal{R}_{j}$ represents the rendered image of the 3DGS model in the $j$-th viewpoint, $\mathcal{F}$ represents the 2D tampering localization backbone, and $\mathbf{M}_{j}$ is the corresponding masks generated by the 2D model in the $j$-th viewpoint. With the masks, we project them back to the 3D space to initially get whether a specific 3D Gaussian is manipulated or not. We define that when the center point's projection falls within the tampered region indicated by the masks, the 3D vote is 1 in that viewpoint. If its projection is within the authentic region, the 3D vote is 0. If the projection lies outside the masks region, the 3D vote is -1, which can be considered as an abstention in that viewpoint.
Specifically, it can be expressed as follows:
\begin{equation}
\small
\mathbf{V}_{i j} =  \mathbb{I}\left[\mu_i \mathbf{P}_j \in \mathbf{M}_j^{+}\right] + 0 \cdot \mathbb{I}\left[\mu_i \mathbf{P}_j \in \mathbf{M}_j^{-}\right] - \mathbb{I}\left[\mu_i \mathbf{P}_j \notin \mathbf{M}_j\right], 
\end{equation}
where $\mathbf{P}_j$ represents the projection matrix in the $j$-th viewpoint, $\mu_i$ represents the coordinates of the $i$-th 3D Gaussian center point. $\mathbf{M}_{j}+$ is the tampered region indicated by the masks, and $\mathbf{M}_{j}-$ is the authentic region.
$\mathbf{V}_{i j}$ represents the value of the $i$-th row and $j$-th column element in the voting matrix $\mathbf{V}$, \textit{i.e.}, it indicates whether the $i$-th 3D Gaussian center point belongs to the manipulated region in the $j$-th viewpoint.

Then, we determine whether each 3D Gaussian belongs to the tampered 3GDS area by calculating its total number of votes in different viewpoints, and the higher number of votes represents the higher probability of its being manipulated. Note that we do not count abstentions in this process. Specifically, it can be calculated as follows:
\begin{equation}
\mathbf{T}_i=\sum_{j=0}^{N-1} \mathbf{V}_{i j},  ~\operatorname{if}  \mu_i \mathbf{P}_j \in \mathbf{M}_j,
\end{equation}
where $N$ represents the number of viewpoints, $\mathbf{T}_i$ represents the total number of votes for the $i$-th 3D Gaussian. 
Then, we treat the 3D Gaussians that receive the majority of votes as being tampered, which implies a consensus in the majority of viewpoints.
This means, the number of viewpoints which consider the 3D Gaussian as tampered regions exceeds that of authentic regions or abstentions.
Since the 2D tampering localization model does not consider the structure in 3D space, it may give inconsistent prediction results in different viewpoints. Therefore, with this 3D voting method, we can utilize the consistency of the 3D structure among different viewpoints to remove some incorrect results. 

Finally, the 3D tampering attribute uses the total number of votes $\mathbf{T}_i$ after reaching consensus as the initial value. We integrate this tampering attribute into the parameters of the 3DGS model. Then, for the $i$-th 3D Gaussian, in addition to the mean position $\boldsymbol{\mu}_i$, opacity $\sigma_i$, scaling factor $\boldsymbol{s}_i$, rotation factor $\boldsymbol{q}_i$ and SH coefficients $\boldsymbol{h}_i$, the 3D tampering attribute $ r_i \in \mathbb{R}^1$ is used to indicate whether this 3D Gaussian has been tampered.

\subsection{3D contrastive mechanism} 
\begin{figure}[t]
  \centering
  \includegraphics[width=\linewidth]{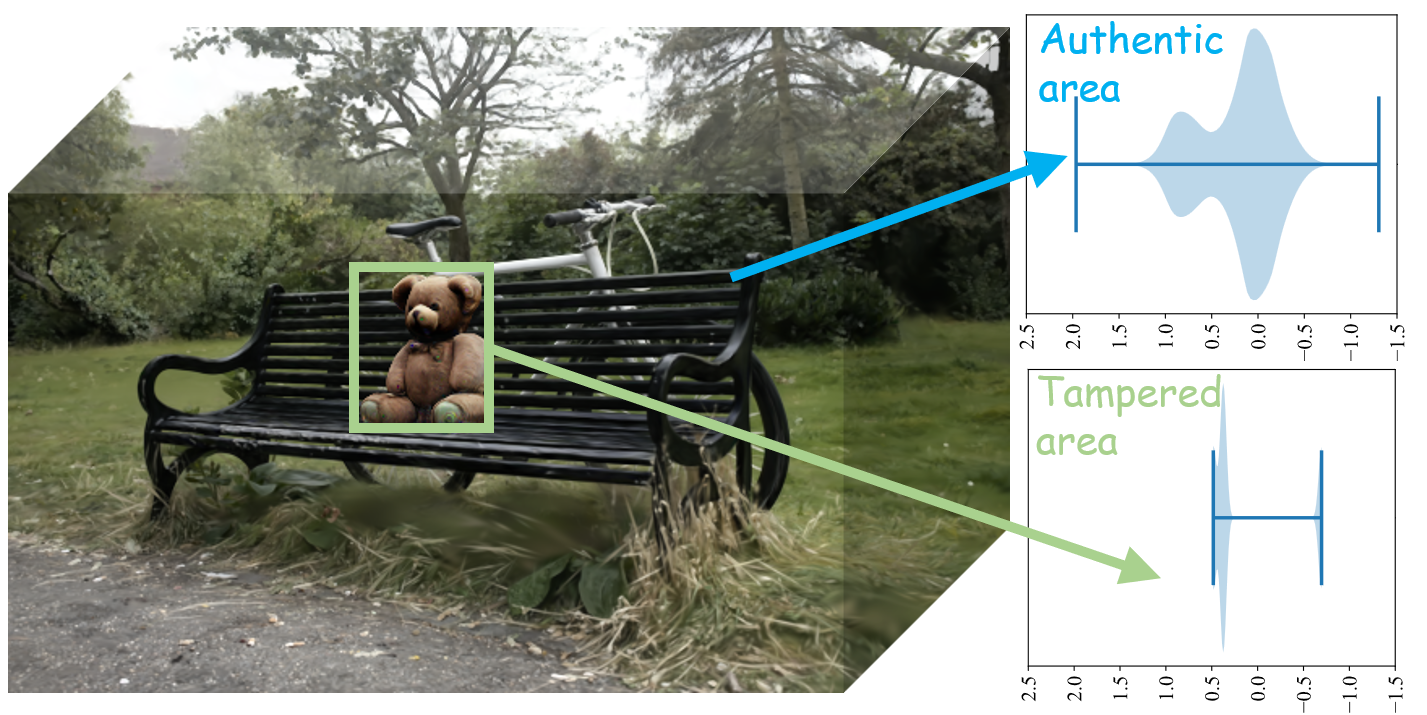}
  \caption{Statistical distribution of tampered and authentic 3D Gaussian properties. Noticeable anomalies are observed in the tampered regions, where parameter distributions deviate from those of the authentic parts. These discrepancies serve as key indicators of tampering, motivating our approach to seek tampering traces at the 3D level.}
  \label{fig:distribution}
\end{figure}

As shown in Figure~\ref{fig:distribution}, we compare the distributions of attribute values (\emph{e.g.,} rotation factor) between tampered and authentic 3D Gaussians. Considering that tampered 3D Gaussians exhibit certain differences in parameter distributions compared with authentic Gaussians, we further leverage the properties of 3D Gaussian to facilitate the performance of tampering localization. 

Therefore, we propose a 3D contrastive mechanism that increases or reduces the corresponding 3D tampering attribute value by comparing the similarity between Gaussian attributes. 
Specifically, we consider Gaussians with high 3D tampering attribute values $\mathcal{G}_h$ as tampered and Gaussians with very low values $\mathcal{G}_l$ as authentic. 
Then the main focus is on those Gaussians with intermediate values $\mathcal{G}_m$. 
The similarity of Gaussian attributes between $\mathcal{G}_m$ and $\mathcal{G}_h$, $\mathcal{G}_l$ is compared to determine whether they are more likely to be tampered or authentic. 
If $\mathcal{G}_m$ is more similar to $\mathcal{G}_h$, the corresponding 3D tampering attribute value is increased, and on the contrary, its 3D tampering attribute value is decreased. The process of comparing Gaussian attributes can be represented as:

\begin{equation}
\begin{aligned}
\operatorname{sim}^h_i &= \|\mathbf{F}^m_i - \mathbf{F}^h\|_2^2, \\
\operatorname{sim}^l_i &= \|\mathbf{F}^m_i - \mathbf{F}^l\|_2^2,
\end{aligned}
\end{equation}
where $\operatorname{sim}^h_i$ represents the similarity between $i$-th 3D Gaussian in $\mathcal{G}_m$ and $\mathcal{G}_h$, $\operatorname{sim}^l_i$ represents the corresponding similarity with $\mathcal{G}_l$. $\mathbf{F}^m_i$ represents $i$-th 3D Gaussian's attributes in $\mathcal{G}_m$. $\mathbf{F}^h$ and $\mathbf{F}^l$ indicate the average value of Gaussian attributes of $\mathcal{G}_h$ and $\mathcal{G}_l$, respectively.
Here, we leverage the original attributes of each 3D Gaussian, including $\boldsymbol{\mu}_i$, $\sigma_i$, $\boldsymbol{s}_i$, $\boldsymbol{q}_i$ and $\boldsymbol{h}_i$. Then, the update direction of 3D tampering attributes can be expressed as:
\begin{equation}
u_i = \operatorname{sign}\left( \Delta \operatorname{sim}_i \right), ~\operatorname{where} ~\Delta \operatorname{sim}_i= \operatorname{sim}_i^{l} - \operatorname{sim}_i^{h},
\end{equation}
where $u_i$ represents the update direction of $i$-th 3D Gaussian's tampering attribute in $\mathcal{G}_m$, $\operatorname{sign}(\cdot)$ indicates the sign function. 
Our 3D contrastive mechanism can be implemented through a loss function, and the detailed description of the formulation is presented in the next section.

\subsection{Cyclic optimization strategy}
After obtaining 3D tampering attributes, we introduce a cyclic optimization strategy to update them iteratively. 
Since the 3D tampering attribute has been integrated as a unique attribute into the 3DGS model parameters, we can render it into 2D by using the differentiable rasterization algorithm. Particularly, we optimize the 3D tampering attribute by calculating the difference between the rendered results and the masks generated by the 2D tampering localization backbone. The process of 3D tampering attribute rendering can be expressed as follows:
\begin{equation}
\mathbf{M}^{R}(\mathbf{r})=\sum_{i=1}^{G}  r_i \alpha_i \prod_{j =1}^{i-1} (1 - \alpha_j),
\end{equation}
where $r_i$ represents the 3D tampering attribute of the $i$-th 3D Gaussian, $G$ represents the number of 3D Gaussians overlapping the ray $\mathbf{r}$, and $\mathbf{M}^{R}(\mathbf{r})$ represents the corresponding rendered results, and \( \alpha_i \) is determined by evaluating a 2D Gaussian with covariance \( \Sigma' \), multiplied by a per-point opacity.

After obtaining the rendered results, we construct a loss function consisting of the rendered results and masks generated by the 2D model to perform the optimization. This allows us to update the 3D tampering attributes by using the gradient descent algorithm. Specifically, our goal is to increase the 3D tampering attribute value of the manipulated area and decrease the value of the authentic area.
Thus, considering that the manipulated area is 1 and authentic area is 0 in the masks, the 2D rendering loss function can be calculated as follows:
\begin{equation}
\begin{aligned}
\mathcal{L}_{2d}= & -\lambda_1\sum_{\mathbf{r} \in \mathrm{R}(\mathbf{I}_j)} \mathbf{M}_j(\mathbf{r}) \mathbf{M}^{R}(\mathbf{r}) \\
& +\lambda_2 \sum_{\mathbf{r} \in \mathrm{R}(\mathbf{I}_j)}\big(1- \mathbf{M}_j(\mathbf{r})\big) \mathbf{M}^{R}(\mathbf{r}),
\end{aligned}
\label{eq10}
\end{equation}
where $\lambda_1$ and $\lambda_2$ are hyperparameters that balance different loss terms, $\mathrm{R}(\mathbf{I}_j)$ indicates the set of rays for image $\mathbf{I}_j$ in the $j$-th viewpoint. In this process, we only optimize the 3D tampering attributes without affecting other parameters of the 3D Gaussian model. 

In our optimization process, we first use 2D rendering loss to optimize the 3D tampering attribute score. 
After a certain number of iterations, the difference of 3D tampering attribute scores between $\mathcal{G}_h$ and $\mathcal{G}_l$ becomes larger to facilitate the distinction. 
Then, we use both 2D rendering loss and 3D contrastive mechanism for optimization. The loss function at this point can be defined as:
\begin{equation}
\mathcal{L}_{cyc} = \beta\mathcal{L}_{2d} + \gamma\mathcal{L}_{3d},
\label{eq11}
\end{equation}
where $\beta$ and $\gamma$ denote hyperparameters that balance 2D rendering loss and 3D loss $\mathcal{L}_{3d}$. The loss function of 3D contrastive mechanism $\mathcal{L}_{3d}$ can be calculated as follows:

\begin{equation}
\mathcal{L}_{3d} = -\frac{1}{K}\sum_{i \in K}u_i r_i,
\end{equation}
where $K$ is the number of 3D Gaussians in $\mathcal{G}_m$. By optimizing this loss function, we can increase or decrease the corresponding 3D tampering attribute scores. After several iterations, we can obtain more accurate results for localizing the manipulated areas in the 3DGS model. 
\subsection{Implementation details}

We implement our method using PyTorch.
As our goal is to locate the tampered areas in 3DGS, we first manipulate the 3DGS models trained in standard settings by using the 3DGS editing methods (\textit{i.e.}, GaussianEditor~\cite{chen2024gaussianeditor} and Gaussian Grouping~\cite{grouping}). Next, the rendered images of these scenes are fed into the 2D tampering localization backbone~\cite{kwon2025safire} to obtain the corresponding masks, where the confidence score threshold is set to 0.5. After obtaining the masks, we employ the 3D voting method to project them back to the 3D space and integrate within each 3D Gaussian.
Then, our approach can update the 3D tampering attributes iteratively, which leverages the 2D rendering loss and 3D contrastive mechanism to perform optimization at both 2D and 3D levels. 
During the optimization process, the weights of the different loss functions in Equation~(\ref{eq10}) and Equation~(\ref{eq11}) are set as $\lambda_1=1.0$, $\lambda_2=10.0$, $\beta=1.0$ and $\gamma=10.0$, and the learning rate for optimizing the 3D tampering attribute values is set to 1.0. We choose the Adam optimizer to update the tampering attributes. The threshold of the 3D tampering attributes after normalization is set to 0.1. All of our experiments are performed on a single NVIDIA Tesla V100 GPU.
\begin{table*}
	\centering
	\begin{NiceTabular}{c | c c | c c | c c}[colortbl-like]
		\toprule
		  \multirow{2}{*}{Method}  & \multicolumn{2}{c|}{Object incorporation} &  \multicolumn{2}{c|}{Object modification}  & \multicolumn{2}{c}{Object removal} \\ 
           & F1$\uparrow$ & IoU$\uparrow$ & F1$\uparrow$ & IoU$\uparrow$  & F1$\uparrow$ & IoU$\uparrow$ \\
		\midrule
		  Inverse with SAFIRE~\cite{kwon2025safire}  & 0.1191 & 0.0647 &  0.2743 & 0.1614 & 0.3896 & 0.2435  \\  
		  SAFIRE~\cite{kwon2025safire}+SA3D~\cite{sa3d} & 0.1576 & 0.0915 & 0.8348 & 0.7168 & 0.0177 & 0.0090 \\
		  SAFIRE~\cite{kwon2025safire}+SAGD~\cite{sagd} & 0.4671 & 0.3880 & 0.6230 & 0.4557 & 0.4866 & 0.3223  \\
		  \textbf{Proposed GS-Checker}  & \textbf{0.9507} & \textbf{0.9081} & \textbf{0.9017} & \textbf{0.8232} & \textbf{0.7812} & \textbf{0.6433} \\
		\bottomrule
	\end{NiceTabular}
         \caption{Comparison of quantitative tampering localization performance with baseline methods. The best results are in \textbf{bold}.}
	\label{tab:edit}
\end{table*}

\begin{figure*}[t]
  \centering
  \includegraphics[width=\linewidth]{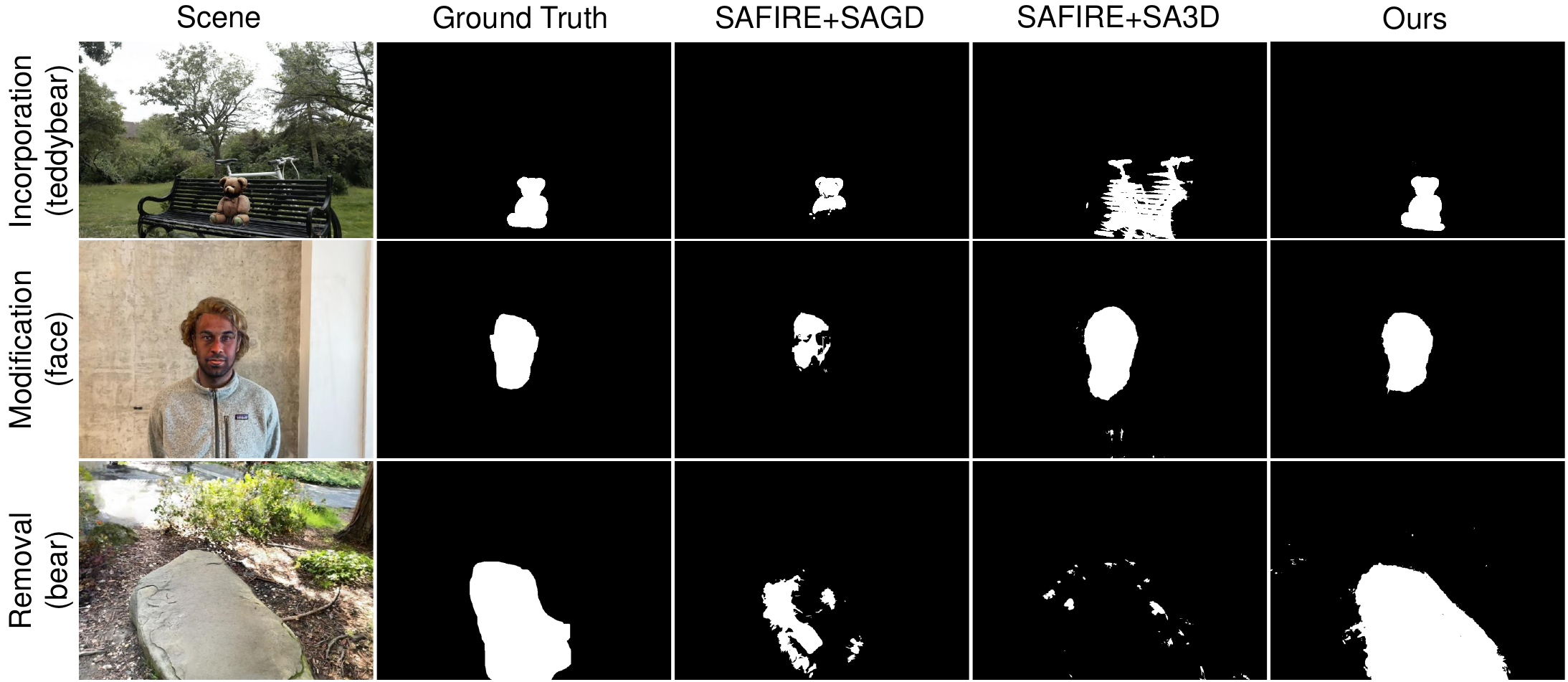}
  \caption{Qualitative results of 3DGS tampering localization in different scenes. Columns from left to right are: rendered images of tampered 3D scenes, ground-truth, SAFIRE~\cite{kwon2025safire}+SAGD~\cite{sagd}, SAFIRE~\cite{kwon2025safire}+SA3D~\cite{sa3d} and ours.}
  \label{fig:qualitative}
\end{figure*}
\section{Experiments}
\subsection{Experimental settings}
\noindent\textbf{Dataset.} In order to evaluate the effectiveness of our proposed method, we construct a 3DGS manipulation dataset containing multiple scenes and manually annotated mask labels. 
Specifically, we use GaussianEditor~\cite{chen2024gaussianeditor} and Gaussian Grouping~\cite{grouping}, two latest and effective 3DGS editing methods, to edit 3DGS models trained on Mip-NeRF360~\cite{barron2022mip} and InstructNeRF2NeRF~\cite{haque2023instruct} datasets. All these 3DGS models are trained under standard settings~\cite{3dgs}. The dataset includes three types of tampering: \textbf{1) Object incorporation:} incorporating objects at a certain location in 3D scenes; \textbf{2) Object modification:} editing the properties of objects in 3D scenes, such as color, shape and so on; \textbf{3) Object removal:} removing some objects from the 3D scenes and filling the holes generated at the interface. In total, our dataset comprises 11 tampered 3DGS models, strictly following the scenes number of 3DGS editing works.

\noindent\textbf{Baselines.} To the best of our knowledge, there is no method specifically for the 3DGS tampering localization. Therefore, we compare with three strategies to guarantee a fair comparison:  \textbf{1) SAFIRE~\cite{kwon2025safire}+SA3D~\cite{sa3d}:} locating 3D tampered regions from the results of 2D tampering localization model~\cite{kwon2025safire} with a state-of-the-art 3DGS segmentation method SA3D~\cite{sa3d}; \textbf{2) SAFIRE~\cite{kwon2025safire}+SAGD~\cite{sagd}:} locating 3D tampered regions from the results of 2D tampering localization model~\cite{kwon2025safire} with a 3DGS segmentation method SAGD~\cite{sagd}; \textbf{3) Inverse with SAFIRE~\cite{kwon2025safire}:} directly project the results generated by the 2D tampering localization model~\cite{kwon2025safire} back to 3D Gaussians.

\noindent\textbf{Evaluation methodology.} We evaluate the proposed method as well as the baseline methods using the standard manipulation localization metrics~\cite{ma2024imdl}. Specifically, we use the average value of $F_1$ score and IoU between the rendered results and ground-truth masks of different viewpoints to evaluate the performance of different methods. 
In addition to the normal situations, we also verify whether the proposed method can retain the 3D tampering localization performance against different distortions, including 2D Gaussian noise, Gaussian blur, 3D scale noise and opacity noise.
\subsection{Experimental results}
\noindent\textbf{Quantitative results.} We compare the tampering localization performance of our method with all baselines on three tampering types, and the results are presented in Table~\ref{tab:edit}.
It can be observed that our method achieves the best tampering localization performance in different tampering types. This proves that our method can efficiently and accurately locate the tampered areas in 3D scenes. 
Comparatively, other methods have lower performance and often struggle to locate the tampered areas in these 3D scenes. 
This discrepancy underscores the challenges associated with directly applying traditional 3DGS segmentation methods or attempting to project masks back to 3D Gaussians. These approaches often fail to identify tampering traces at the 3D level.
In contrast, our proposed method can be well applied to the 3DGS tampering localization task, which demonstrates the effectiveness of our method for 3DGS tampering localization.

\noindent\textbf{Qualitative results.} We visualize the qualitative results of 3DGS tampering localization in different scenes, and the results are shown in Figure~\ref{fig:qualitative}. 
It can be observed that our proposed method can locate the tampered areas effectively across different 3D scenes. This demonstrates the effectiveness of our method.
In comparison, the other methods are relatively inferior in localizing the tampered areas.

\begin{table}[t]
	\centering
        \begin{NiceTabular}{c | c c c c}[colortbl-like]
		\toprule
            \multirow{2}{*}{Distortion}   &  \multicolumn{2}{c}{FIRE+3D} &  \multicolumn{2}{c}{Ours}  \\
            ~ & F1$\uparrow$ & IoU$\uparrow$ & F1$\uparrow$ & IoU$\uparrow$\\
		\midrule
            None & 0.8348 & 0.7168 & 0.9017 & 0.8232\\
		  Gauss. noise & 0.4806 & 0.3176 & 0.8903 & 0.8046 \\  
		  Gauss. blur  & 0.8088 &  0.6795 & 0.8568 & 0.7532 \\
		  scale noise & 0.8302 & 0.7100 & 0.8801 & 0.7877  \\  
		  opacity noise & 0.8129 & 0.6851 & 0.8959 & 0.8140 \\
		\bottomrule
	\end{NiceTabular}
        \caption{Quantitative tampering localization performance of our method and SAFIRE~\cite{kwon2025safire}+SA3D~\cite{sa3d} (FIRE+3D) under different distortions. ``None" indicates that no distortion has been applied.}
	\label{tab:robust}
\end{table}

\noindent\textbf{Robustness evaluation.} To verify the robustness of the proposed method, we perform 3DGS tampering localization experiments under different distortions, including 2D Gaussian noise and Gaussian blur. 
We also verify the tampering localization performance in the case of adding noise to 3D Gaussian parameters, including adding noise to the scaling parameters as well as the opacity parameters of the 3DGS model.
The results of the robustness evaluation experiments are shown in Table~\ref{tab:robust}. It can be observed that our method can still maintain a relatively accurate 3D tampering localization performance under different distortions, thus avoiding being misguided to produce false results.

\begin{figure}[t]
  \centering
  \includegraphics[width=\linewidth]{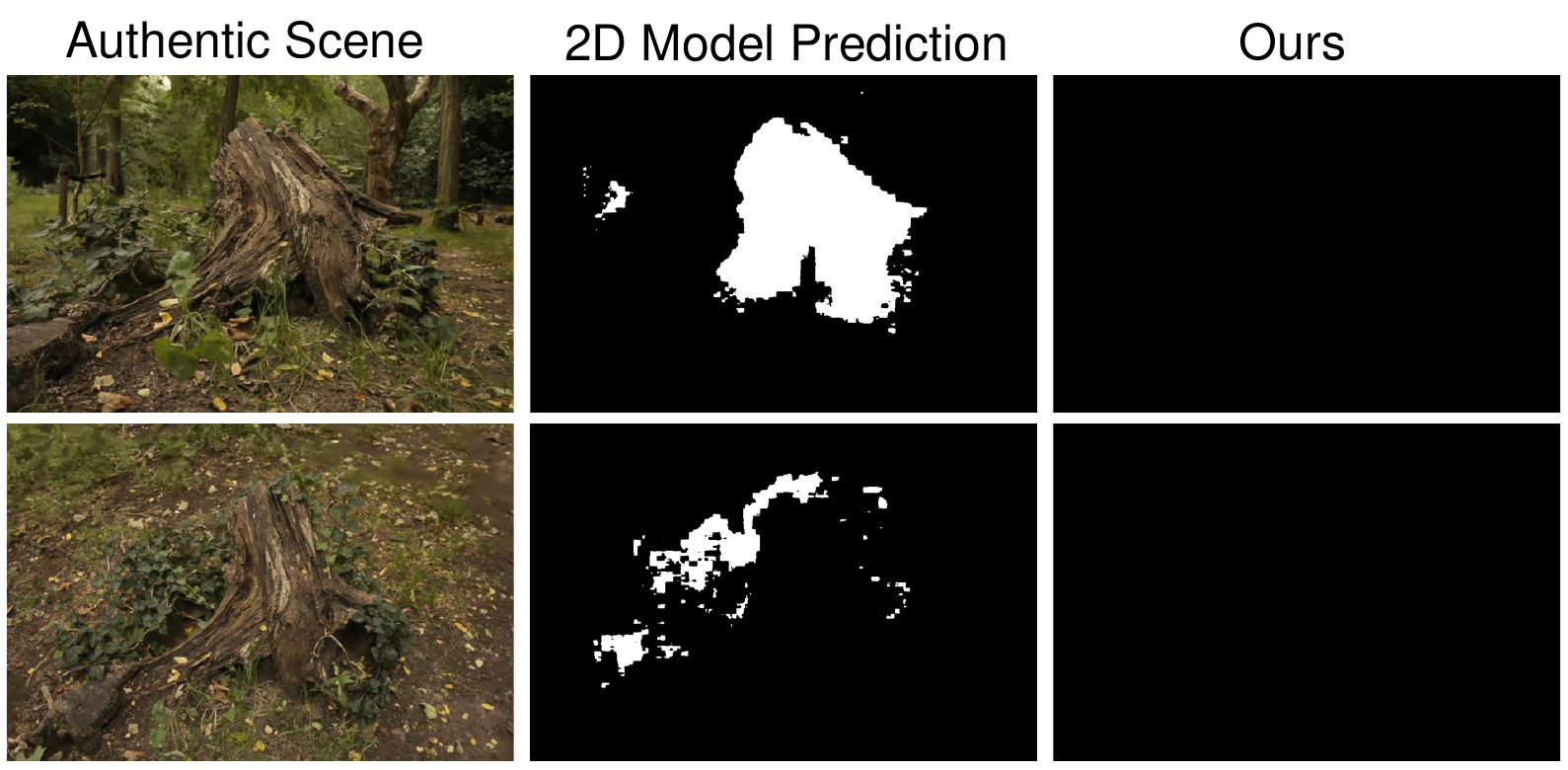}
  \caption{Results on the authentic scene \textit{stump}.}
  \label{fig:authentic}
\end{figure}

\noindent\textbf{Results on authentic scene.} We conduct experiments on the authentic scene to verify whether our proposed method would incorrectly detect the authentic scene as tampered or not. The experimental results are shown in Figure~\ref{fig:authentic}. It can be observed that the 2D tampering localization model incorrectly predicts some areas of the authentic scene as tampered, while our method can remove these erroneously localized areas. 
This proves the effectiveness of our method and the credibility of its prediction results.

\begin{figure}[t]
  \centering
  \includegraphics[width=\linewidth]{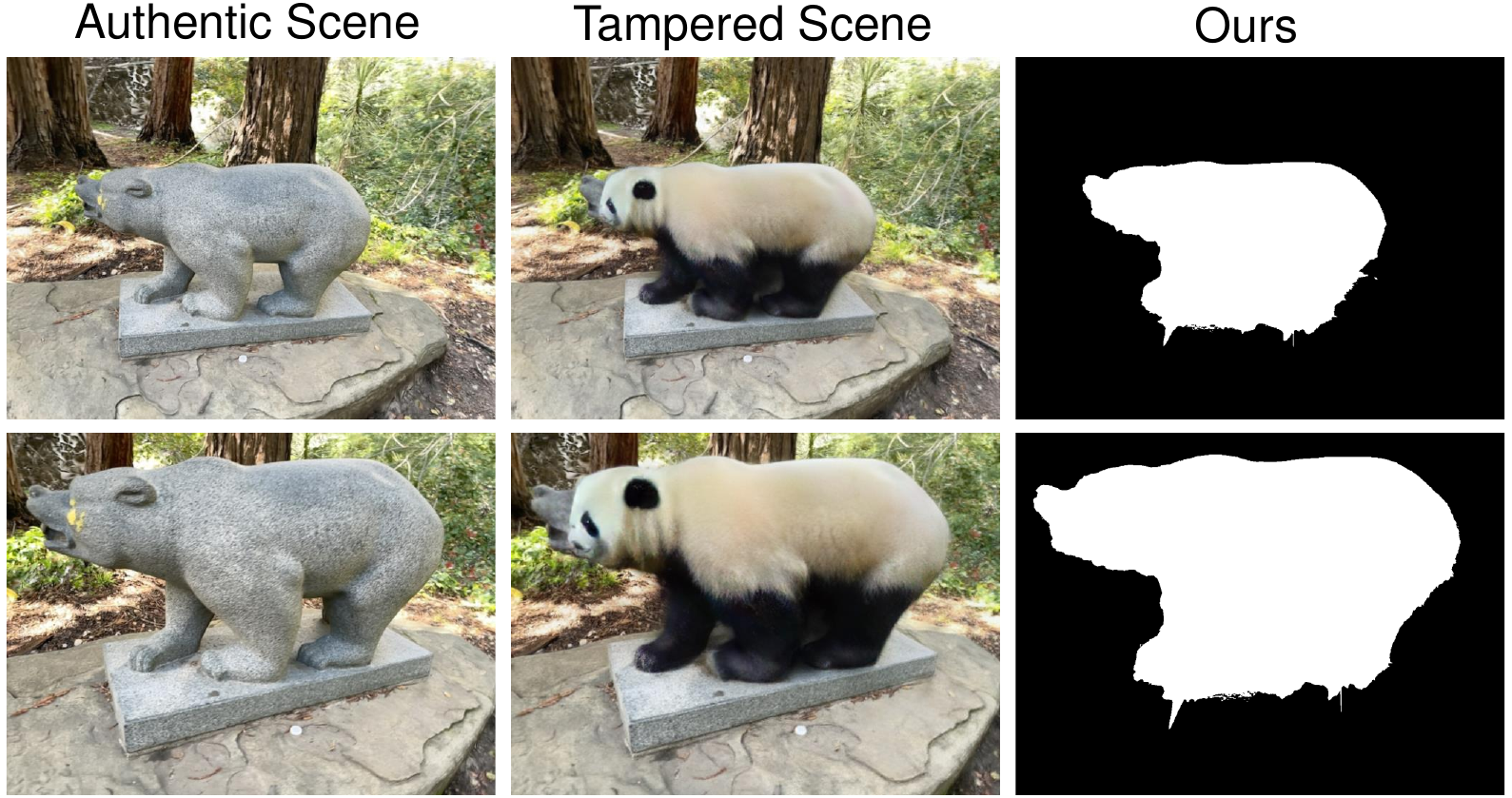}
  \caption{Qualitative tampering localization results on DGE~\cite{chen2024dge} editing method.}
  \label{fig:editor}
\end{figure}

\noindent\textbf{Results on different 3DGS editing methods.} 
In addition to GaussianEditor~\cite{chen2024gaussianeditor} and Gaussian Grouping~\cite{grouping}, we also evaluate the performance of our method on other editing methods, such as DGE~\cite{chen2024dge}.
The experimental results are presented in Figure~\ref{fig:editor}. It can be found that our method has excellent tampering localization performance on different 3DGS editing methods. This demonstrates the effectiveness and generalization of our method.

\begin{figure}[t]
  \centering
  \includegraphics[width=\linewidth]{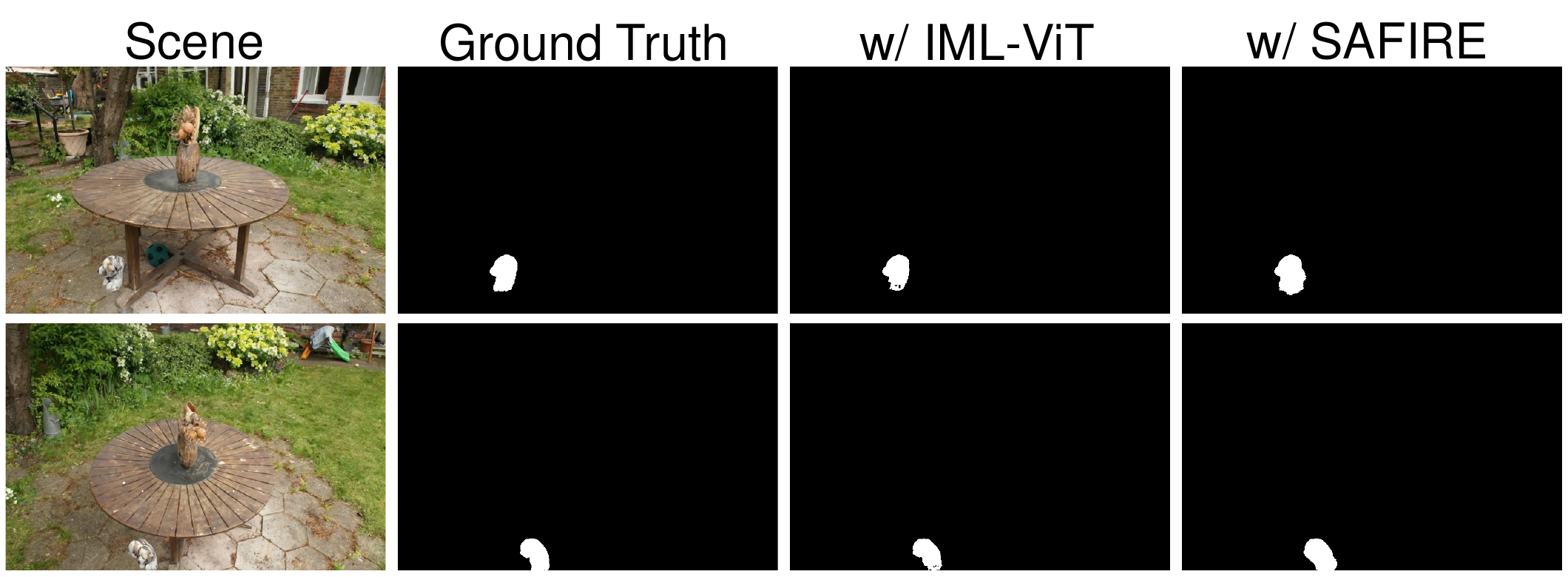}
  \caption{Qualitative tampering localization results on different 2D models. Our method is flexible and capable of integrating with different 2D tampering localization models to ensure accurate 3DGS tampering localization.}
  \label{fig:base_model}
\end{figure}

\noindent\textbf{Different 2D tampering localization models.} We conduct experiments on different 2D tampering localization models as well, including IML-ViT~\cite{ma2023iml} and SAFIRE~\cite{kwon2025safire}. The experimental results are presented in Figure~\ref{fig:base_model}. 
It can be observed that our approach is adaptable and can incorporate various 2D tampering localization models to achieve precise 3DGS tampering localization. 
This flexibility enhances the robustness and effectiveness of our proposed method.

\noindent\textbf{Ablation study.} In this section, we further perform ablation studies to verify the effectiveness of 3D contrastive mechanism and cyclic optimization strategy.
Table~\ref{abla} shows the results of different component combinations for locating 3DGS tampered areas. 
It can be observed that each component has an improvement on the performance of the 3DGS tampering localization, which demonstrates the effectiveness of each component.

\begin{table}
	\centering
    \begin{NiceTabular}{c c | c c}[colortbl-like]
		\toprule
		  Case & Method & F1$\uparrow$ & IoU$\uparrow$    \\
		\midrule
		  (a) & w/o cyclic optimization & 0.8485 & 0.7384 \\
            (b) & w/o 3D contrastive  & 0.8788 & 0.7847  \\ 
		(c) & Our full method & \textbf{0.8842} & \textbf{0.7934}  \\
		\bottomrule
	\end{NiceTabular}
    \caption{Comparisons for our full method, our method without cyclic optimization strategy and our method without 3D contrastive mechanism.}
\label{abla}
\end{table}

\section{Conclusion}
In this paper, we introduce \textbf{\textit{GS-Checker}}, a novel approach to locate the tampered areas of 3D Gaussian Splatting (3DGS) models. 
By integrating a 3D tampering attribute into 3D Gaussian parameters, we can exploit the properties
of the 3DGS model for capturing tampering information effectively.
Additionally, a 3D contrastive mechanism is employed to identify tampering cues by comparing the similarity of key attributes among 3D Gaussians.
We further adopt a cyclic optimization strategy to iteratively refine the tampering attribute, achieving precise localization results. 
Extensive experimental results show that our method can achieve accurate 3DGS tampering localization performance and be robust to different distortions.
Thus, it can effectively prevent 3DGS models from being used by malicious users for negative applications.

\section*{Acknowledgement}
Renjie Group is supported by the National Natural Science Foundation of China under Grant No. 62302415, Guangdong Basic and Applied Basic Research Foundation under Grant No.  2024A1515012822, and the Research Grant Council (RGC) of the Hong Kong SAR, under a GRF Grant 12203124 and a ECS Grant 22201125. We also thank the support of the São Paulo Research Foundation (Fapesp) through the Horus Project \#2023/12865-8 and the Brazilian National Council for Scientific and Technological Research (CNPq) through complementary grants.

\bibliography{aaai2026}

\begin{thebibliography}{47}
\providecommand{\natexlab}[1]{#1}

\bibitem[{Bao et~al.(2023)Bao, Zhang, Yang, Fan, Yang, Bao, Zhang, and Cui}]{bao2023sine}
Bao, C.; Zhang, Y.; Yang, B.; Fan, T.; Yang, Z.; Bao, H.; Zhang, G.; and Cui, Z. 2023.
\newblock Sine: Semantic-driven image-based nerf editing with prior-guided editing field.
\newblock In \emph{Proceedings of the IEEE/CVF Conference on Computer Vision and Pattern Recognition}.

\bibitem[{Barron et~al.(2022)Barron, Mildenhall, Verbin, Srinivasan, and Hedman}]{barron2022mip}
Barron, J.~T.; Mildenhall, B.; Verbin, D.; Srinivasan, P.~P.; and Hedman, P. 2022.
\newblock Mip-{N}e{RF} 360: Unbounded anti-aliased neural radiance fields.
\newblock In \emph{Proceedings of the IEEE/CVF Conference on Computer Vision and Pattern Recognition}.

\bibitem[{Cen et~al.(2025)Cen, Fang, Zhou, Yang, Xie, Zhang, Shen, and Tian}]{sa3d}
Cen, J.; Fang, J.; Zhou, Z.; Yang, C.; Xie, L.; Zhang, X.; Shen, W.; and Tian, Q. 2025.
\newblock Segment anything in 3d with radiance fields.
\newblock \emph{International Journal of Computer Vision}.

\bibitem[{Chen, Laina, and Vedaldi(2024)}]{chen2024dge}
Chen, M.; Laina, I.; and Vedaldi, A. 2024.
\newblock {DGE}: Direct gaussian 3{D} editing by consistent multi-view editing.
\newblock In \emph{European Conference on Computer Vision}.

\bibitem[{Chen et~al.(2024{\natexlab{a}})Chen, Chen, Zhang, Wang, Yang, Wang, Cai, Yang, Liu, and Lin}]{chen2024gaussianeditor}
Chen, Y.; Chen, Z.; Zhang, C.; Wang, F.; Yang, X.; Wang, Y.; Cai, Z.; Yang, L.; Liu, H.; and Lin, G. 2024{\natexlab{a}}.
\newblock Gaussian{E}ditor: Swift and controllable 3{D} editing with gaussian splatting.
\newblock In \emph{Proceedings of the IEEE/CVF Conference on Computer Vision and Pattern Recognition}.

\bibitem[{Chen et~al.(2024{\natexlab{b}})Chen, Zhang, Yang, Cai, Yu, Yang, and Lin}]{chen2024it3d}
Chen, Y.; Zhang, C.; Yang, X.; Cai, Z.; Yu, G.; Yang, L.; and Lin, G. 2024{\natexlab{b}}.
\newblock I{T3D}: Improved text-to-3{D} generation with explicit view synthesis.
\newblock In \emph{Proceedings of the AAAI Conference on Artificial Intelligence}.

\bibitem[{Dong et~al.(2022)Dong, Chen, Hu, Cao, and Li}]{dong2022mvss}
Dong, C.; Chen, X.; Hu, R.; Cao, J.; and Li, X. 2022.
\newblock Mvss-net: Multi-view multi-scale supervised networks for image manipulation detection.
\newblock \emph{IEEE Transactions on Pattern Analysis and Machine Intelligence}.

\bibitem[{Gao et~al.(2023)Gao, Aigerman, Groueix, Kim, and Hanocka}]{gao2023textdeformer}
Gao, W.; Aigerman, N.; Groueix, T.; Kim, V.; and Hanocka, R. 2023.
\newblock Textdeformer: Geometry manipulation using text guidance.
\newblock In \emph{ACM SIGGRAPH 2023 Conference Proceedings}.

\bibitem[{Guillaro et~al.(2023)Guillaro, Cozzolino, Sud, Dufour, and Verdoliva}]{guillaro2023trufor}
Guillaro, F.; Cozzolino, D.; Sud, A.; Dufour, N.; and Verdoliva, L. 2023.
\newblock Trufor: Leveraging all-round clues for trustworthy image forgery detection and localization.
\newblock In \emph{Proceedings of the IEEE/CVF conference on computer vision and pattern recognition}.

\bibitem[{Guo et~al.(2023)Guo, Liu, Ren, Grosz, Masi, and Liu}]{guo2023hierarchical}
Guo, X.; Liu, X.; Ren, Z.; Grosz, S.; Masi, I.; and Liu, X. 2023.
\newblock Hierarchical fine-grained image forgery detection and localization.
\newblock In \emph{Proceedings of the IEEE/CVF Conference on Computer Vision and Pattern Recognition}.

\bibitem[{Haque et~al.(2023)Haque, Tancik, Efros, Holynski, and Kanazawa}]{haque2023instruct}
Haque, A.; Tancik, M.; Efros, A.~A.; Holynski, A.; and Kanazawa, A. 2023.
\newblock Instruct-{N}e{RF}2{N}e{RF}: Editing 3{D} scenes with instructions.
\newblock In \emph{Proceedings of the IEEE/CVF International Conference on Computer Vision}.

\bibitem[{Hu et~al.(2024)Hu, Wang, Fan, Fan, Peng, Lei, Li, and Zhang}]{sagd}
Hu, X.; Wang, Y.; Fan, L.; Fan, J.; Peng, J.; Lei, Z.; Li, Q.; and Zhang, Z. 2024.
\newblock {SAGD}: Boundary-enhanced segment anything in 3D Gaussian via Gaussian decomposition.
\newblock arXiv:2401.17857.

\bibitem[{Huang et~al.(2024)Huang, Li, Cheung, Cheung, See, and Wan}]{huang2024gaussianmarker}
Huang, X.; Li, R.; Cheung, Y.-m.; Cheung, K.~C.; See, S.; and Wan, R. 2024.
\newblock Gaussian{M}arker: Uncertainty-aware copyright protection of 3{D} gaussian splatting.
\newblock \emph{Advances in Neural Information Processing Systems}.

\bibitem[{Huang et~al.(2025)Huang, Luo, Song, Wang, and Wan}]{huang2025marksplatter}
Huang, X.; Luo, Z.; Song, Q.; Wang, R.; and Wan, R. 2025.
\newblock MarkSplatter: Generalizable watermarking for {3D} gaussian splatting model via splatter image structure.
\newblock In \emph{Proceedings of the 33rd ACM International Conference on Multimedia}.

\bibitem[{Islam et~al.(2020)Islam, Long, Basharat, and Hoogs}]{islam2020doa}
Islam, A.; Long, C.; Basharat, A.; and Hoogs, A. 2020.
\newblock DOA-GAN: Dual-order attentive generative adversarial network for image copy-move forgery detection and localization.
\newblock In \emph{Proceedings of the IEEE/CVF conference on computer vision and pattern recognition}.

\bibitem[{Kerbl et~al.(2023)Kerbl, Kopanas, Leimk{\"u}hler, and Drettakis}]{3dgs}
Kerbl, B.; Kopanas, G.; Leimk{\"u}hler, T.; and Drettakis, G. 2023.
\newblock 3D Gaussian Splatting for Real-Time Radiance Field Rendering.
\newblock \emph{ACM Transactions on Graphics}.

\bibitem[{Kopanas et~al.(2022)Kopanas, Leimk{\"u}hler, Rainer, Jambon, and Drettakis}]{kopanas2022neural}
Kopanas, G.; Leimk{\"u}hler, T.; Rainer, G.; Jambon, C.; and Drettakis, G. 2022.
\newblock Neural point catacaustics for novel-view synthesis of reflections.
\newblock \emph{ACM Transactions on Graphics}.

\bibitem[{Kopanas et~al.(2021)Kopanas, Philip, Leimk{\"u}hler, and Drettakis}]{kopanas2021point}
Kopanas, G.; Philip, J.; Leimk{\"u}hler, T.; and Drettakis, G. 2021.
\newblock Point-Based Neural Rendering with Per-View Optimization.
\newblock In \emph{Computer Graphics Forum}.

\bibitem[{Kwon et~al.(2025)Kwon, Lee, Nam, Son, and Kim}]{kwon2025safire}
Kwon, M.-J.; Lee, W.; Nam, S.-H.; Son, M.; and Kim, C. 2025.
\newblock {SAFIRE}: Segment any forged image region.
\newblock In \emph{Proceedings of the AAAI Conference on Artificial Intelligence}.

\bibitem[{Li and Huang(2019)}]{li2019localization}
Li, H.; and Huang, J. 2019.
\newblock Localization of deep inpainting using high-pass fully convolutional network.
\newblock In \emph{proceedings of the IEEE/CVF international conference on computer vision}.

\bibitem[{Li and Cheung(2024)}]{li2024variational}
Li, R.; and Cheung, Y.-m. 2024.
\newblock Variational multi-scale representation for estimating uncertainty in 3{D} gaussian splatting.
\newblock \emph{Advances in Neural Information Processing Systems}.

\bibitem[{Li and Cheung(2025)}]{li2025modeling}
Li, R.; and Cheung, Y.-m. 2025.
\newblock Modeling and Identifying Distractors with Curriculum for Robust 3D Gaussian Splatting.
\newblock In \emph{Proceedings of the 33rd ACM International Conference on Multimedia}.

\bibitem[{Li and Zhou(2018)}]{li2018fast}
Li, Y.; and Zhou, J. 2018.
\newblock Fast and effective image copy-move forgery detection via hierarchical feature point matching.
\newblock \emph{IEEE Transactions on Information Forensics and Security}.

\bibitem[{Liu et~al.(2021)Liu, Zhang, Zhang, Zhang, Zhu, and Russell}]{liu2021editing}
Liu, S.; Zhang, X.; Zhang, Z.; Zhang, R.; Zhu, J.-Y.; and Russell, B. 2021.
\newblock Editing conditional radiance fields.
\newblock In \emph{Proceedings of the IEEE/CVF international conference on computer vision}.

\bibitem[{Luo et~al.(2025)Luo, Rocha, Shi, Guo, Li, and Wan}]{luo2025nerf}
Luo, Z.; Rocha, A.; Shi, B.; Guo, Q.; Li, H.; and Wan, R. 2025.
\newblock The {N}e{RF} signature: Codebook-aided watermarking for neural radiance fields.
\newblock \emph{IEEE Transactions on Pattern Analysis and Machine Intelligence}.

\bibitem[{Ma et~al.(2023)Ma, Du, Jiang, Hammadi, and Zhou}]{ma2023iml}
Ma, X.; Du, B.; Jiang, Z.; Hammadi, A. Y.~A.; and Zhou, J. 2023.
\newblock {IML-ViT}: Benchmarking image manipulation localization by vision transformer.
\newblock arXiv:2307.14863.

\bibitem[{Ma et~al.(2024)Ma, Zhu, Su, Du, Jiang, Tong, Lei, Yang, Pun, Lv et~al.}]{ma2024imdl}
Ma, X.; Zhu, X.; Su, L.; Du, B.; Jiang, Z.; Tong, B.; Lei, Z.; Yang, X.; Pun, C.-M.; Lv, J.; et~al. 2024.
\newblock IMDL-BenCo: A Comprehensive Benchmark and Codebase for Image Manipulation Detection \& Localization.
\newblock In \emph{Advances in Neural Information Processing Systems}.

\bibitem[{Mikaeili et~al.(2023)Mikaeili, Perel, Safaee, Cohen-Or, and Mahdavi-Amiri}]{mikaeili2023sked}
Mikaeili, A.; Perel, O.; Safaee, M.; Cohen-Or, D.; and Mahdavi-Amiri, A. 2023.
\newblock Sked: Sketch-guided text-based 3{D} editing.
\newblock In \emph{Proceedings of the IEEE/CVF International Conference on Computer Vision}.

\bibitem[{Poole et~al.(2022)Poole, Jain, Barron, and Mildenhall}]{poole2022dreamfusion}
Poole, B.; Jain, A.; Barron, J.~T.; and Mildenhall, B. 2022.
\newblock Dreamfusion: Text-to-3{D} using 2{D} diffusion.
\newblock In \emph{International Conference on Learning Representations}.

\bibitem[{Radford et~al.(2021)Radford, Kim, Hallacy, Ramesh, Goh, Agarwal, Sastry, Askell, Mishkin, Clark et~al.}]{radford2021learning}
Radford, A.; Kim, J.~W.; Hallacy, C.; Ramesh, A.; Goh, G.; Agarwal, S.; Sastry, G.; Askell, A.; Mishkin, P.; Clark, J.; et~al. 2021.
\newblock Learning transferable visual models from natural language supervision.
\newblock In \emph{International conference on machine learning}.

\bibitem[{Raj et~al.(2023)Raj, Kaza, Poole, Niemeyer, Ruiz, Mildenhall, Zada, Aberman, Rubinstein, Barron et~al.}]{raj2023dreambooth3d}
Raj, A.; Kaza, S.; Poole, B.; Niemeyer, M.; Ruiz, N.; Mildenhall, B.; Zada, S.; Aberman, K.; Rubinstein, M.; Barron, J.; et~al. 2023.
\newblock Dreambooth3{D}: Subject-driven text-to-3{D} generation.
\newblock In \emph{Proceedings of the IEEE/CVF international conference on computer vision}.

\bibitem[{Rombach et~al.(2022)Rombach, Blattmann, Lorenz, Esser, and Ommer}]{rombach2022high}
Rombach, R.; Blattmann, A.; Lorenz, D.; Esser, P.; and Ommer, B. 2022.
\newblock High-resolution image synthesis with latent diffusion models.
\newblock In \emph{Proceedings of the IEEE/CVF conference on computer vision and pattern recognition}.

\bibitem[{Sella et~al.(2023)Sella, Fiebelman, Hedman, and Averbuch-Elor}]{sella2023vox}
Sella, E.; Fiebelman, G.; Hedman, P.; and Averbuch-Elor, H. 2023.
\newblock Vox-e: Text-guided voxel editing of 3{D} objects.
\newblock In \emph{Proceedings of the IEEE/CVF International Conference on Computer Vision}.

\bibitem[{Shao et~al.(2023)Shao, Sun, Peng, Zheng, Zhou, Zhang, and Liu}]{shao2023control4d}
Shao, R.; Sun, J.; Peng, C.; Zheng, Z.; Zhou, B.; Zhang, H.; and Liu, Y. 2023.
\newblock Control4d: Dynamic portrait editing by learning 4d gan from 2d diffusion-based editor.
\newblock arXiv:2305.20082.

\bibitem[{Snavely, Seitz, and Szeliski(2006)}]{snavely2006photo}
Snavely, N.; Seitz, S.~M.; and Szeliski, R. 2006.
\newblock Photo tourism: exploring photo collections in 3{D}.
\newblock In \emph{ACM Transactions on Graphics}.

\bibitem[{Song et~al.(2024{\natexlab{a}})Song, Luo, Cheung, See, and Wan}]{song2024geometry}
Song, Q.; Luo, Z.; Cheung, K.~C.; See, S.; and Wan, R. 2024{\natexlab{a}}.
\newblock Geometry cloak: Preventing tgs-based 3{D} reconstruction from copyrighted images.
\newblock \emph{Advances in Neural Information Processing Systems}.

\bibitem[{Song et~al.(2024{\natexlab{b}})Song, Luo, Cheung, See, and Wan}]{song2024protecting}
Song, Q.; Luo, Z.; Cheung, K.~C.; See, S.; and Wan, R. 2024{\natexlab{b}}.
\newblock Protecting {N}e{RF}s’ copyright via plug-and-play watermarking base model.
\newblock In \emph{European Conference on Computer Vision}.

\bibitem[{Song et~al.(2025)Song, Luo, Cheung, See, and Wan}]{song2025align}
Song, Q.; Luo, Z.; Cheung, K.~C.; See, S.; and Wan, R. 2025.
\newblock Align 3D Representation and Text Embedding for 3D Content Personalization.
\newblock In \emph{Proceedings of the 33rd ACM International Conference on Multimedia}.

\bibitem[{Wang et~al.(2022{\natexlab{a}})Wang, Chai, He, Chen, and Liao}]{wang2022clip}
Wang, C.; Chai, M.; He, M.; Chen, D.; and Liao, J. 2022{\natexlab{a}}.
\newblock Clip-{N}e{RF}: Text-and-image driven manipulation of neural radiance fields.
\newblock In \emph{Proceedings of the IEEE/CVF Conference on Computer Vision and Pattern Recognition}.

\bibitem[{Wang et~al.(2023)Wang, Jiang, Chai, He, Chen, and Liao}]{wang2023nerf}
Wang, C.; Jiang, R.; Chai, M.; He, M.; Chen, D.; and Liao, J. 2023.
\newblock {N}e{RF}-{A}rt: Text-driven neural radiance fields stylization.
\newblock \emph{IEEE Transactions on Visualization and Computer Graphics}.

\bibitem[{Wang et~al.(2022{\natexlab{b}})Wang, Wu, Chen, Han, Shrivastava, Lim, and Jiang}]{wang2022objectformer}
Wang, J.; Wu, Z.; Chen, J.; Han, X.; Shrivastava, A.; Lim, S.-N.; and Jiang, Y.-G. 2022{\natexlab{b}}.
\newblock Objectformer for image manipulation detection and localization.
\newblock In \emph{Proceedings of the IEEE/CVF Conference on Computer Vision and Pattern Recognition}.

\bibitem[{Wu et~al.(2024)Wu, Bian, Li, Wang, Reid, Torr, and Prisacariu}]{wu2024gaussctrl}
Wu, J.; Bian, J.-W.; Li, X.; Wang, G.; Reid, I.; Torr, P.; and Prisacariu, V.~A. 2024.
\newblock Gaussctrl: Multi-view consistent text-driven 3{D} gaussian splatting editing.
\newblock In \emph{European Conference on Computer Vision}.

\bibitem[{Xu and Harada(2022)}]{xu2022deforming}
Xu, T.; and Harada, T. 2022.
\newblock Deforming radiance fields with cages.
\newblock In \emph{European Conference on Computer Vision}.

\bibitem[{Ye et~al.(2024)Ye, Danelljan, Yu, and Ke}]{grouping}
Ye, M.; Danelljan, M.; Yu, F.; and Ke, L. 2024.
\newblock Gaussian {G}rouping: Segment and edit anything in 3{D} scenes.
\newblock In \emph{European Conference on Computer Vision}.

\bibitem[{Zhang et~al.(2024{\natexlab{a}})Zhang, Li, Yu, Xu, Li, and Zhang}]{zhang2024editguard}
Zhang, X.; Li, R.; Yu, J.; Xu, Y.; Li, W.; and Zhang, J. 2024{\natexlab{a}}.
\newblock Editguard: Versatile image watermarking for tamper localization and copyright protection.
\newblock In \emph{Proceedings of the IEEE/CVF Conference on Computer Vision and Pattern Recognition}.

\bibitem[{Zhang et~al.(2024{\natexlab{b}})Zhang, Meng, Li, Xu, Zhang, and Zhang}]{zhang2024gs}
Zhang, X.; Meng, J.; Li, R.; Xu, Z.; Zhang, Y.; and Zhang, J. 2024{\natexlab{b}}.
\newblock {GS-H}ider: Hiding messages into 3{D} gaussian splatting.
\newblock \emph{Advances in Neural Information Processing Systems}.

\bibitem[{Zhang et~al.(2025)Zhang, Tang, Xu, Li, Xu, Chen, Gao, and Zhang}]{zhang2025omniguard}
Zhang, X.; Tang, Z.; Xu, Z.; Li, R.; Xu, Y.; Chen, B.; Gao, F.; and Zhang, J. 2025.
\newblock Omniguard: Hybrid manipulation localization via augmented versatile deep image watermarking.
\newblock In \emph{Proceedings of the Computer Vision and Pattern Recognition Conference}.

\end{thebibliography}

\end{document}